\documentclass[10pt,twocolumn,letterpaper]{article}

\usepackage[pagenumbers]{cvpr} 

\usepackage{graphicx}
\usepackage{amsmath}
\usepackage{amssymb}
\usepackage{booktabs}
\usepackage{rotating}

\usepackage{color,colortbl}
\usepackage{multirow}
\usepackage{multicol}
\usepackage{pifont}         

\definecolor{Gray}{gray}{0.9}

\newlength\savewidth\newcommand\shline{\noalign{\global\savewidth\arrayrulewidth
		\global\arrayrulewidth 1pt}\hline\noalign{\global\arrayrulewidth\savewidth}}

\usepackage[pagebackref,breaklinks,colorlinks]{hyperref}

\usepackage[capitalize]{cleveref}
\crefname{section}{Sec.}{Secs.}
\Crefname{section}{Section}{Sections}
\Crefname{table}{Table}{Tables}
\crefname{table}{Tab.}{Tabs.}

\def\cvprPaperID{4} 
\def\confName{CVPR}
\def\confYear{2022}

\begin{document}

\title{CORE: Consistent Representation Learning for Face Forgery Detection}

\author{
Yunsheng Ni$^{1}$
~
Depu Meng$^{2}$
~
Changqian Yu$^{3}$
~
Chengbin Quan$^{1}$
~
Dongchun Ren$^{3}$
~
Youjian Zhao$^{1}$\thanks{Corresponding author.
This work was done when Y. Ni, D. Meng
were interns at Meituan, Beijing, China.} \\
\normalsize
$^{1}$ Tsinghua University
~~
$^{2}$ University of Science and Technology of China
~~
$^{3}$ Meituan \\
\small\texttt{nys19@mails.tsinghua.edu.cn}
~~
\texttt{mdp@mail.ustc.edu.cn} \\
\small\texttt{\{yuchangqian, rendongchun\}@meituan.com}
~~
\texttt{\{quancb, zhaoyoujian\}@tsinghua.edu.cn}
}

\maketitle

\begin{abstract}

Face manipulation techniques develop rapidly and arouse widespread public concerns.
Despite that vanilla convolutional neural networks achieve acceptable performance,
they suffer from the overfitting issue.
To relieve this issue,
there is a trend to introduce some erasing-based augmentations.
We find that these methods indeed attempt to
implicitly induce more consistent representations for different augmentations 
via assigning the same label for different augmented images.
However, due to the lack of explicit regularization, 
the consistency between different representations is less satisfactory.
Therefore, 
we constrain the consistency of different representations explicitly
and propose a simple yet effective framework, COnsistent REpresentation Learning~(CORE).
Specifically, we first capture the different representations with different augmentations,
then regularize the cosine distance of the representations
to enhance the consistency.
Extensive experiments (in-dataset and cross-dataset) demonstrate that
CORE performs favorably against state-of-the-art face forgery detection methods.
Our code is available at \href{https://github.com/niyunsheng/CORE}{https://github.com/niyunsheng/CORE}.

\end{abstract}

\section{Introduction}
\label{sec:intro}

With the rapid development of 
face manipulation techniques (\eg autoencoder-based~\cite{li2020celeb} and GAN-based~\cite{korshunova2017fast, nirkin2019fsgan,karras2019style}), 
synthetic faces become extremely hard for humans to distinguish from real faces.
This causes a considerable
risk to the trust and security of society.
Thus, it is important to develop effective methods for face forgery detection.

Since a vanilla convolutional neural network~(CNN)
has achieved acceptable performance,
it suffers from the overfitting issue~\cite{li2020face, wang2021representative, sohrawardi2019poster,qian2020thinking,liu2021spatial,chen2021local,das2021towards} 
due to oversampling the real faces to generate the fake samples.
To relieve this issue,
recent works~\cite{wang2021representative, das2021towards}
introduce some effective erasing-based data augmentations.
They erase different regions of a sample to capture more general forgery representation.
In fact,
through observing their class activation mapping~(CAM),
we find that the erasing-based methods 
tend to \textbf{implicitly} induce more \textbf{consistent} representations 
for the sample of different data augmentations. 
Nevertheless, due to the lack of explicit regularization, 
the consistency between different representations is less satisfactory.
Therefore,
we \textbf{explicitly} constrain the representation distances between different data augmentations
to capture more intrinsic forgery evidence.

\begin{figure}[t]
    \centering
    \includegraphics[width=.95\columnwidth]{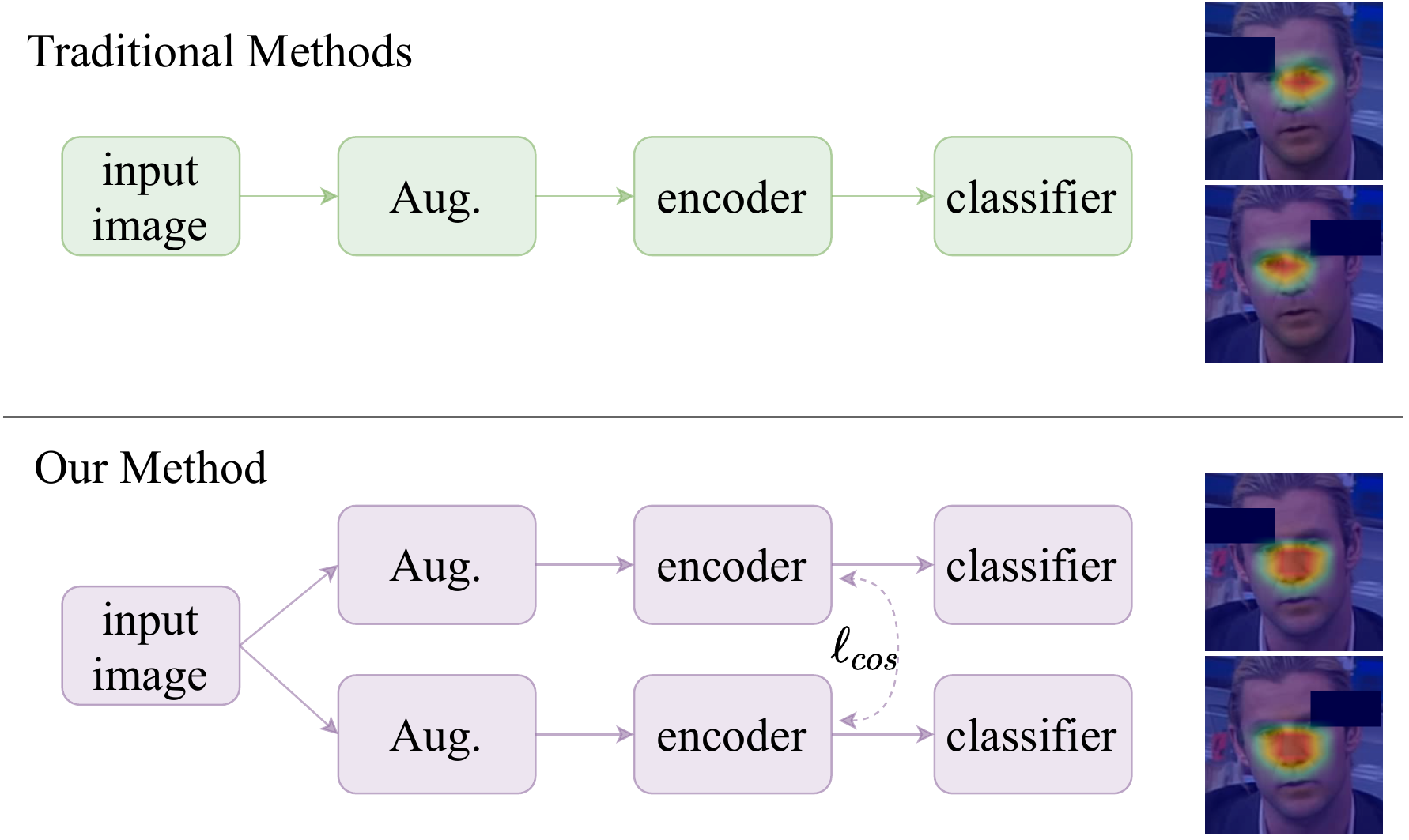}
    \caption{An illustration of the CORE framework compared
    to traditional methods. 
    The right column shows the Class Activation Mapping (CAM) of input of erasing different parts. The CAM is more consistent in our method.
    Aug. refers to data augmentations.}
    \label{fig:framework}
\end{figure}

In this paper, we propose a simple yet effective framework, Consistent Representation Learning (CORE), to explicitly learn consistent representations for face forgery detection.
As shown in Fig.~\ref{fig:framework},
the proposed framework adopts paired random augmentations
to transform the input to different views.
A shared Encoder extracts the corresponding feature 
representations
for the transformed inputs.
We explicitly constrain the consistency of the different representations via a Consistency Loss.
Finally, a Classifier Network assigns the supervised label
for each representation.

Compared with the traditional methods, 
the proposed framework has two merits:
(1) The representations of different augmentations are regularized explicitly, which enables the model to attend to more intrinsic forgery evidence. 
(2) Our framework does not modify the model structure and can be flexibly integrated with almost any other method.

The main contributions of our work are as follows:
\begin{enumerate}
    \item We propose a simple yet effective framework, Consistent Representation Learning~(CORE). This framework captures different representations of the same sample via different augmentations, and constrains them consistent explicitly with the Consistency Loss.
    \item The proposed framework enables a vanilla CNN model to obtain state-of-the-art performance on FF++~\cite{rossler2019faceforensics++}(RAW, HQ), Celeb-DF~\cite{li2020celeb}, and DFFD~\cite{dang2020detection} benchmarks for in-dataset evaluation, on DFD~\cite{dufour2019contributing} for cross-dataset evaluation.
\end{enumerate}
\section{Related Work}

\noindent\textbf{Face forgery detection.} 
Most recent works formulate face forgery detection as a binary classification problem. 
FaceForensics++ (FF++)~\cite{rossler2019faceforensics++}
proposes a benchmark and provide a baseline
with a vanilla Xception~\cite{chollet2017xception} network. 
Patch~\cite{chai2020makes} regards the input image as some patches and uses a shallow network as the backbone.
Face X-ray~\cite{li2020face} aims to localize the blending boundary in a self-supervised mechanism. 
Multi-attention~\cite{zhao2021multi} proposes a novel multi-attentional network architecture.
Two-branch RN~\cite{masi2020two} learns representations combining the color domain and the frequency domain.
F3-Net~\cite{qian2020thinking} and RFAM~\cite{chen2021local} mine clues in the frequency domain,
and achieve impressive performance in low-quality videos.
LSC~\cite{zhao2021learning} hypothesizes that images’
distinct source features can be preserved and 
extracted after going through deepfake generation processes.
Thus the inconsistency of source features can be used
to detect deepfakes generations. 
Our proposed method CORE tries to learn
consistent representation for face forgery detection
that is invariant to data augmentations.

\begin{figure*}[t]
    \centering
    \includegraphics[width=0.95\textwidth]{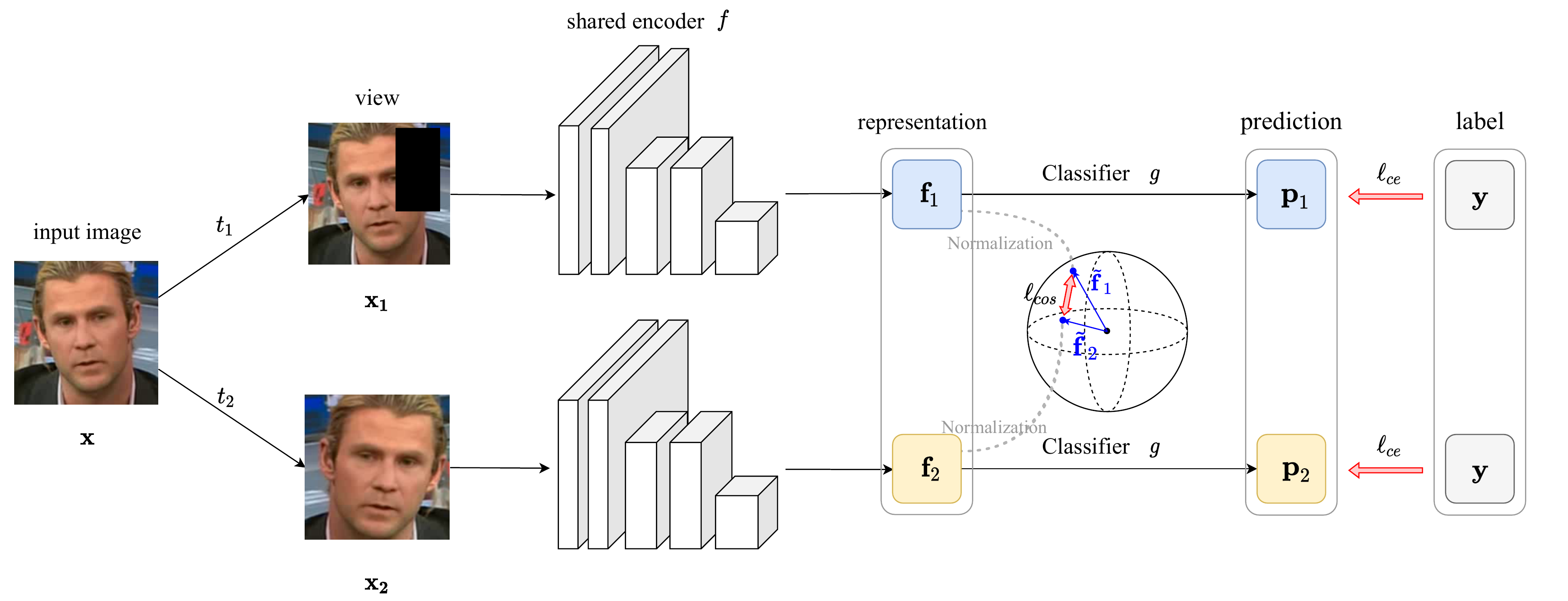}
    
    \caption{An illustration of the architecture of Consistent Representation Learning.
    We first adopt two different data augmentations to the input image $\mathbf{x}$ and get two views $\mathbf{x_1}$ and $\mathbf{x_2}$. Then we feed the two views into a shared encoder $f$ and get two representations $\mathbf{f_1}$ and $\mathbf{f_2}$. 
    We feed the two representations into a classifier $g$ and use a
    cross-entropy loss to train the network, 
    as well as adopt a cosine similarity loss $\ell_{cos}$ 
    to penalize the distance between $\mathbf{f_1}$ and $\mathbf{f_2}$.
    }
    \label{fig:arch}
\end{figure*}

\noindent\textbf{Data augmentation.}
Data augmentations are proven to be useful in computer vision
problems. 
Random Erasing~\cite{zhong2020random} randomly erases a rectangle area of the input.
Adversarial Erasing~\cite{wei2017object} is initially used
in the weakly supervised semantic segmentation,
which erases the activate areas produced by Class Activation Mapping (CAM)~\cite{zhou2016learning}.
Some works study the data augmentations in face forgery
detection~\cite{bondi2020training, charitidis2020face, wang2021representative,das2021towards}. 
RFM~\cite{wang2021representative} traces the facial region which is sensitive to the network and erases the top-N areas.
Face-Cutout~\cite{das2021towards} uses the facial landmarks and randomly cuts out the face part (eg. mouth, eye, etc.)
Data augmentations help to learn
representations invariant to certain transformations.
We explicitly force the model to learn certain invariance by
a consistency loss.

\noindent\textbf{Consistency learning.}
Consistency learning,
enforcing the predictions or features of 
different views for the same unlabeled instance
to be similar,
has been widely applied in
semi-supervised learning, such as $\Pi$ Model~\cite{LaineA17},
Temporal Ensemble~\cite{LaineA17}, and Mean Teacher~\cite{TarvainenV17}.
We show that learning consistent representations
is also helpful for face forgery detection under
a fully-supervised setting.

\noindent\textbf{Contrastive learning.}
Contrastive learning achieves great success in
unsupervised visual representation learning.
MoCo~\cite{he2020momentum}, SimCLR~\cite{ChenK0H20},
InstDisc~\cite{WuXYL18} build instance-level
contrastive learning framework.
PixPro~\cite{XieL00L021}, DenseCL~\cite{WangZSKL21}
build pixel-wise contrastive learning framework.
There are some recent works~\cite{ cozzolino2021towards,fung2021deepfakeucl,xu2022supervised}
introduce contrastive learning into face forgery detection
and get a considerable generalization performance.
These methods adopt popular instance discrimination
based or pixel wise contrastive learning
framework for face forgery detection.
We do not apply instance-level or pixel-level
contrastive learning.
Instead, we only use consistency regularization
to learn representation that attend to extract more intrinsic
forgery evidence.

\section{Proposed Method}

\subsection{Preliminaries}

Face forgery detection is often formulated as a binary classification problem.
The face forgery detection pipeline follows the standard
image classification pipeline:
data pre-processing (including data augmentation),
feature extraction, and prediction through a classifier.
As shown in Fig.~\ref{fig:arch},
we generate two views of the same image by applying random data augmentation twice,
and penalize the distance between the representations of the two views.

\subsection{Consistent Representation Learning Framework}

Given an input batch of $N$ images, we first apply random data augmentations twice to obtain $2N$ images ($N$ pairs).
Then we put the augmented images to an encoder and get $2N$ representation vectors. 
The representation vectors are fed
into a classifier to obtain classification scores.
Moreover, 
we apply a consistency loss on
the representation vectors from a pair of images
for explicitly learning augmentation invariance.

As illustrated in Fig.~\ref{fig:arch}, 
our framework is composed of three components:
data augmentation, encoder network, and classifier
network.

\noindent\textbf{Data augmentation.} 
Given a set of transformations $\mathcal{T}$
and an input image $\mathbf{x}$,
two transformations $t_1, t_2$ are randomly sampled from
$\mathcal{T}$ to serve as data augmentations.
Two views of the image are produced through
the two augmentations: 
$\mathbf{x}_1 = t_1(\mathbf{x})$,
$\mathbf{x}_2 = t_2(\mathbf{x})$.
The data augmentations should not 
alter the intrinsic property of its genuineness.
See Sec.~\ref{sec:implementation} 
for more details about data augmentations.

\noindent\textbf{Encoder network.} 
We use a Xception\cite{chollet2017xception} network
as our encoder network $f$. 
The encoder network maps the two views of input image
$\mathbf{x}_1, \mathbf{x}_2$ into two
$d-$dimensional representation vectors
$\mathbf{f}_1 = f(\mathbf{x}_1), \mathbf{f}_2 = f(\mathbf{x}_2)$.
The two representation vectors are fed into a classifier
network for classification, as well as 
for consistency loss computation.

\noindent\textbf{Classifier network.}
The classifier network (donate as $g$) contains a linear layer and a softmax normalization layer, 
mapping a representation vector $\mathbf{f}$ 
into a scalar probability $p$
(we adopt the second dimension of the output after softmax layer as final probability),
$p = g(\mathbf{f})$. 
The output probability is used for fake face classification.

\subsection{Loss Functions}

\noindent\textbf{Consistency loss.}
Given this framework, then we introduce the consistency loss.
The consistency loss is used to 
penalize the distances of the 
representation vectors that are
extracted from different views
of the same original image. 

We adopt cosine similarity loss ($\ell_{cos}$) to penalize the distance between the two representation vectors, as follows:
\begin{align}
\ell_{cos}\left(\mathbf{f}^{(n)}_{1}, \mathbf{f}^{(n)}_{2}\right)  
            = \left(1-\tilde{\mathbf{f}}^{(n)}_{1} \cdot \tilde{\mathbf{f}}^{(n)}_{2}\right)^{2}       
\end{align}
where $\tilde{\mathbf{f}}=\frac{\mathbf{f}}{\|\mathbf{f}\|_{2}}$ denotes the normalized vector of the representation vector $\mathbf{f}$. 
As illustrated in Fig.~\ref{fig:arch}, 
feature used for the similarity computation
is normalized by a L$2$ normalization layer firstly.
For $N$ pairs of input images,
the consistency loss can be written as
\begin{align}
\mathcal{L}_{c}= \sum_{n=1}^{N} \ell_{cos}\left(\mathbf{f}^{(n)}_{1}, \mathbf{f}^{(n)}_{2}\right)=\sum_{n=1}^{N} \left(1-\tilde{\mathbf{f}}^{(n)}_{1} \cdot \tilde{\mathbf{f}}^{(n)}_{2}\right)^{2}
\end{align}

Cosine similarity loss only pulls the angle of the
vectors to be similar,
ignoring the norm of the vectors.
The reason we choose to use cosine
similarity is that we do not
force the representations of different
views to be exactly the same.
As shown in Fig.~\ref{fig:arch},
the RE augmentation can cut out
a region in the face, which
makes the information in the two
views not identical.
In this case, forcing the representation
to be the same might harm to
the learned representations.
Thus, we use cosine loss
instead of L1 or L2 loss
that also forces the norm
of two vectors to be the same.
Empirical analysis on different consistency
losses is given in Sec.~\ref{sec:ablation}.

\noindent\textbf{Classification loss.}
We use standard cross-entropy loss as classification loss:
\begin{align}
\ell_{ce}(p)=y \log {p}+(1-y) \log (1-p)
\end{align}
where $y$ is the ground-truth label.
For $N$ pairs of input images, the classification loss can be written as
\begin{align}
    \mathcal{L}_{ce}=\sum_{n=1}^{N} \left( \ell_{ce}\left(p^{(n)}_{1}\right) + \ell_{ce}\left(p^{(n)}_{2}\right)\right)
\end{align}

\noindent\textbf{Overall loss.}
We combine the consistency loss with the cross-entropy loss
to form the overall loss in our framework:
\begin{align}
    \mathcal{L}=\mathcal{L}_{ce}+\alpha \mathcal{L}_{c}
\end{align}
Here $\alpha$ is a balance weight for the two losses.

\section{Experiments}

\subsection{Dataset}

We conduct extensive experiments on five well-known datasets: FaceForensics++ (FF++)~\cite{rossler2019faceforensics++}, Celeb-DF~\cite{li2020celeb}, DFFD~\cite{dang2020detection}, DFDC Preview (DFDC-P)~\cite{dolhansky2019deepfake} and deepfakeDetection (DFD)~\cite{dufour2019contributing}. 
FF++ is a widely used face forgery dataset. 
There are a total  of $1000$ real videos ($720$ videos for training, $140$ videos for validation, $140$ videos for testing) and $4,000$ manipulated videos generated with four forgery methods.
Celeb-DF contains $5,939$ high-quality manipulated videos and $590$ real video clips collected from YouTube. Using improved synthesis process forgery faces in Celeb-DF are more realistic with fewer traces of forgery visible to human eyes. 
DFFD is more diverse in manipulated types compared to other datasets. There are $58,703$ real and $240,336$ fake still images along with $1,000$ real and $3,000$ fake video clips. Especially, the Deepfacelab subset in DFFD is not available, so we evaluate the metrics following~\cite{wang2021representative} without the Deepfacelab subset.
DFDC-P are mainly low-quality videos and diverse in several axes (gender, skin-tone, age, etc.). There are $4,119$ manipulated videos and $1,131$ real videos.
DFD is released as a complement to the FF++ dataset, 
which contains $363$ real videos and $3,068$ fake videos.

\subsection{Implementation Details}
\label{sec:implementation}

\noindent\textbf{Data pre-processing.} For each video in Celeb-DF, we extract $3$ frames per second. 
For each video in FF++, we sample $270$ frames per video following~\cite{rossler2019faceforensics++}.
For videos in DFD and DFDC-P, we random select $50$ frames per video.
We crop the faces with bounding boxes (detected boxes enlarged $1.3\times$) which is provided by MTCNN~\cite{zhang2016joint}. 
Some errors that crop real faces in manipulated videos occur whether select the biggest or highest detection probability face by MTCNN, 
especially when there are two characters in a video. 
We solved the problem by using the provided video mask in FF++.

\noindent\textbf{Data augmentation.} 
For data augmentation, we use
two basic transformations:  Random Erasing (RE)~\cite{zhong2020random},
and Random Resized Crop (RandCrop)
and two complex augmentation strategies.
For RE transformation, we use scale factor
$(0.02, 0.2)$ and aspect ratio $(0.5, 2)$.
For RandCrop transformation, scale factor
$(1/1.3, 1)$ and aspect ratio $(0.9, 1.1)$
are adopted.
For each input image, there is $1/3$ probability
that the image is not augmented, $1/3$
probability that RE is applied,
$1/3$ probability that RandCrop is applied.
We denote this data augmentation strategy
as RaAug.
Additionally, we use another complex data augmentation contains quality compression, Gauss noise, Gauss blur, random shift, random scale, see~\cite{selim2020} for more details.
We denote this data augmentation strategy
as DFDC\_selim.

\noindent\textbf{Training}. All models are trained using Adam~\cite{kingma2014adam} with a constant learning rate of $0.0002$. We set the input size $299 \times 299$ and mini-batch $32$. All of our models use Xception~\cite{chollet2017xception} as the backbone. All the models are trained within $30$ epochs and with early stopping if no gains are observed in consecutive $5$ epochs. 
We adopt a weight of $(4,1)$ for cross-entropy loss in all experiments to reduce the implicate of less real data. 
ImageNet~\cite{deng2009imagenet} pretrained weights are used
as parameter initialization.

\noindent\textbf{Evaluation.} 
We report Area Under Curve (AUC) as the main metric 
and accuracy (Acc.) as the secondary metric.
For in-dataset experiments, 
we adopt
True Detect Rate (TDR) at False Detect Rate (FDR) of $0.01\%$ (denoted as TDR$_{0.01\%}$) and $0.1\%$ (denoted as TDR$_{0.1\%}$) as a supplementary following~\cite{dang2020detection}.

\begin{table}[t]
    \centering
    \setlength{\tabcolsep}{8pt}
        \small
            \renewcommand{\arraystretch}{1.3}
    \caption{In-dataset comparison between consistent representation learning
    to Xception, and Xception+ (Xception trained with
    Random Erasing data augmentation).
    Our approach performs better than Xception+,
    which verifies the effectiveness of proposed
    consistent representation learning.}
    \begin{tabular}{l|cccc}
        \shline
        Method & AUC & TDR$_{0.1\%}$ & TDR$_{0.01\%}$ \\
        \shline
        Xception & $99.923$ & $92.869$ & $83.764$  \\ 
        Xception+ & $99.930$ & $91.806$ & $86.318$  \\ 
        \hline
        Ours & $\mathbf{99.943}$ & $\mathbf{94.142}$ & $\mathbf{88.388}$ \\
        \shline
    \end{tabular}
    
    \label{tab:results-base}
\end{table}

\begin{table}[t]
    \centering
    \setlength{\tabcolsep}{9.5pt}
        \small
            \renewcommand{\arraystretch}{1.3}
    \caption{In-dataset ablation study on different penalties
    for consistency loss. 
    Cosine penalty performs the best.}
    \begin{tabular}{l|cccc}
        \shline
        Penalty & AUC & TDR$_{0.1\%}$ & TDR$_{0.01\%}$ \\
        \shline
        $-$ & $99.930$ & $93.059$ & $87.239$  \\ 
        L$1$ & $99.938$ & $93.295$ & $84.653$  \\ 
        L$2$ & $99.930$ & $91.377$ & $83.286$  \\ 
        Cos. & $\mathbf{99.943}$ & $\mathbf{94.142}$ & $\mathbf{88.388}$ \\ 
        \shline 
    \end{tabular}
    
    \label{tab:ablation-loss}
\end{table}

\begin{figure*}[t]
    \centering
    \begin{turn}{90} 
    \footnotesize
    ~~Forgery Region
    \end{turn}\hspace{.45mm}
    \includegraphics[width=0.115\textwidth]{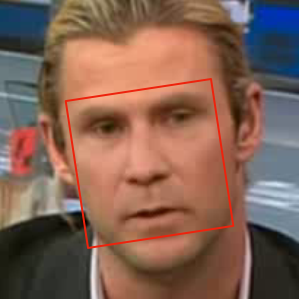}
    \includegraphics[width=0.115\textwidth]{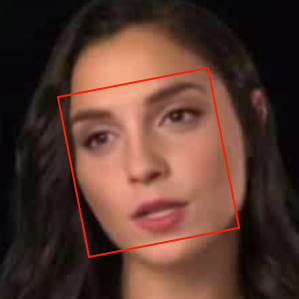}
    \includegraphics[width=0.115\textwidth]{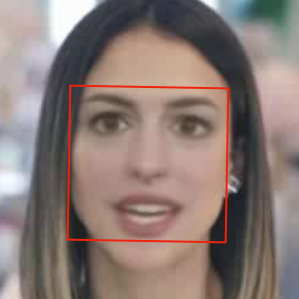}
    \includegraphics[width=0.115\textwidth]{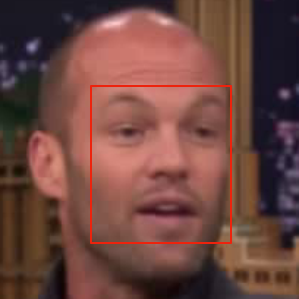}
    \includegraphics[width=0.115\textwidth]{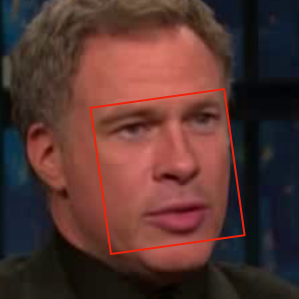}
    \includegraphics[width=0.115\textwidth]{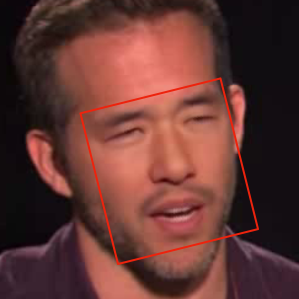}
    \includegraphics[width=0.115\textwidth]{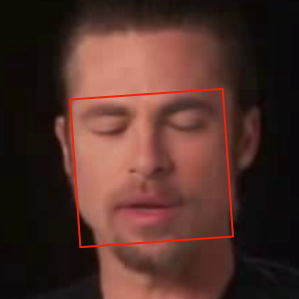}
    \includegraphics[width=0.115\textwidth]{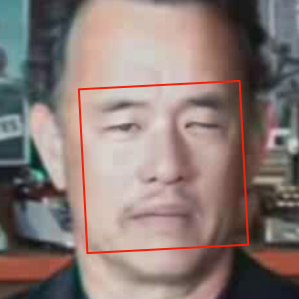}
    \\
    \vspace{.5mm}
    \begin{turn}{90} 
    \footnotesize
    ~~~~~~~~~CAM
    \end{turn}\hspace{1mm}
    \includegraphics[width=0.115\textwidth]{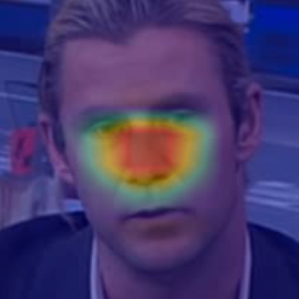}
    \includegraphics[width=0.115\textwidth]{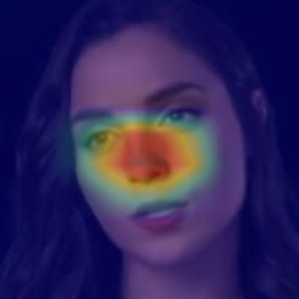}
    \includegraphics[width=0.115\textwidth]{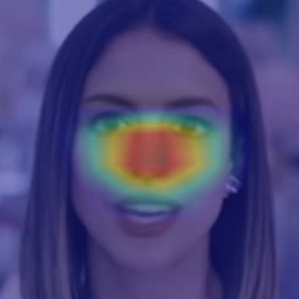}
    \includegraphics[width=0.115\textwidth]{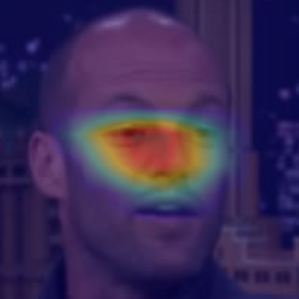}
    \includegraphics[width=0.115\textwidth]{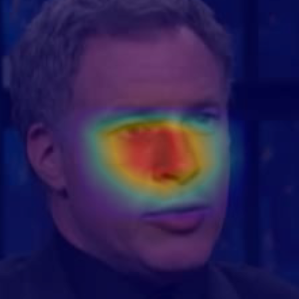}
    \includegraphics[width=0.115\textwidth]{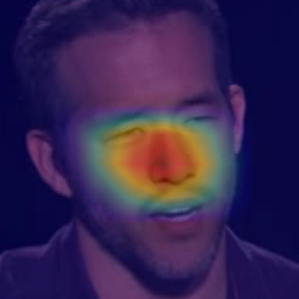}
    \includegraphics[width=0.115\textwidth]{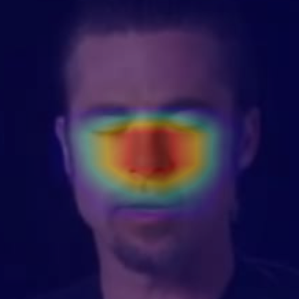}
    \includegraphics[width=0.115\textwidth]{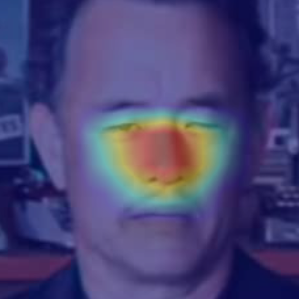}
    \\
    \caption{Visualizations of Class Activation Mapping (CAM)
    of CORE. 
    The CAM highlights an area inside the forgery region. }
    \label{fig:vis-cam}
\end{figure*}

\subsection{In-dataset Ablation Study}
\label{sec:ablation}

In this section, we validate the effectiveness of our proposed
consistent representation learning approach and study two key components in our
method: consistency loss and data augmentation.
We uses Xception as backbone in all experiments.
For in-dataset ablation study, we conduct experiments on Celeb-DF dataset.
We use Random Erasing data augmentation and balance weight $\alpha=1$ as default. 

\noindent\textbf{Effectiveness of consistent representation learning.} 
We compare our consistent representation learning approach with two baselines: Xception and Xception+. Xception refers to an Xception network trained without data augmentation.
Xception+ refers to Xception trained with Random Erasing augmentation. Our approach adopts the same RE data augmentation as Xception+.
As shown in Tab.~\ref{tab:results-base},
Xception+ performs better than Xception, and
our approach performs best, outperforming Xception+ by 
$0.013$ AUC, 
$2.336$ TDR$_{0.1\%}$,
and $2.070$ TDR$_{0.01\%}$.

\noindent\textbf{Ablation study on consistency loss.} 
We study three penalties for consistency loss:
L1, L2, and cosine penalty.
As shown in Tab.~\ref{tab:ablation-loss},
model with L1 or L2 penalty performs
on par or better
than baseline for AUC, while worse
for TDR$_{0.01\%}$.
Model with cosine penalty performs better
than baseline on all AUC, TDR$_{0.1\%}$,
and TDR$_{0.01\%}$ metrics.
We guess the reason for the unsatisfactory
performance of L1 and L2 penalty might be
that: forcing representations of two views
to be identical is harmful to feature learning.

\begin{table}[t]
    \centering
    \setlength{\tabcolsep}{3.8pt}
        \small
            \renewcommand{\arraystretch}{1.3}
    \caption{In-dataset ablation study on balance weight $\alpha$. 
    Model with $\alpha=2$ performs the best.}
    \begin{tabular}{l|c|c|c|c|c|c}
        \shline
        \multicolumn{1}{c|}{$\alpha$} & $0$ & $1$ & $2$ & $5$ & $10$ & $100$ \\
        \shline
        AUC & $99.930$ & $99.943$ & $\mathbf{99.960}$ & $99.943$ & $99.948$ & $99.915$ \\
        \shline
    \end{tabular}
    \label{tab:ablation-weight}
\end{table}

\begin{table}[t]
    \centering
    \setlength{\tabcolsep}{5pt}
        \small
            \renewcommand{\arraystretch}{1.3}
    \caption{In-dataset ablation study on data augmentation. 
    Our approach performs consistently better
    than baseline on all augmentations.
    Ours with RaAug performs the best. }
    \begin{tabular}{l|l|ccc}
        \shline
        Data Aug. & Method & AUC & TDR$_{0.1\%}$ & TDR$_{0.01\%}$ \\
        \shline
        None & Baseline & $99.923$ & $92.869$ & $83.764$  \\ 
        \hline
        \multirow{2}{*}{RE} &  Baseline & $99.930$ & $91.806$ & $86.318$  \\ 
        & Ours & $99.960$ & $\mathbf{95.670}$ & $90.562$ \\ 
        \hline
        \multirow{2}{*}{RFM} &  Baseline & $99.926$ & $91.768$ & $84.419$  \\ 
        & Ours & $99.947$ & $95.014$ & $93.742$ \\ 
        \hline
        \multirow{2}{*}{RandCrop} & Baseline & $99.929$ & $89.432$ & $84.504$  \\ 
        & Ours & $99.945$ & $91.823$ & $84.065$ \\
        \hline
        \multirow{2}{*}{RaAug} & Baseline & $99.909$ & $90.553$ & $85.615$  \\
        & Ours & $\mathbf{99.971}$ & $95.575$ & $\mathbf{93.980}$ \\
        \hline
        \multirow{2}{*}{DFDC\_selim} & Baseline & $99.895$ & $90.256$ & $83.732$\\ 
        & Ours & $99.941$ & $93.970$ & $88.737$\\
        \shline
    \end{tabular}
    
    \label{tab:results-dataaug}
\end{table}

\noindent\textbf{Ablation study on balance weight $\alpha$.}
We explore the balance weight setting for consistency loss in Tab.~\ref{tab:ablation-weight}. In all cases except $\alpha=100$, consistent representation learning improves the performance of the model trained without consistency loss, and we find that
model with $\alpha = 2$ achieves the best performance.

\noindent\textbf{Ablation study on data augmentation.} 
We conduct experiments on different data
augmentation strategies:
RE~\cite{zhong2020random}, RFM~\cite{wang2021representative}, RandCrop, and 
proposed RaAug (see Sec.~\ref{sec:implementation}
for details).
We use Xception models trained with the
same data augmentation as baselines
and balance weight $\alpha=2$.
As shown in Tab.~\ref{tab:results-dataaug},
our model consistently
outperforms baseline under all data augmentations,
by $+0.02$ to $+0.07$ AUC improvement
and $+2.4$ to $+5.0$ TDR$_{0.1\%}$ improvement.
Especially, RaAug performs overall the best among
four data augmentation strategies, with $99.971$ AUC
and $93.980$ TDR$_{0.01\%}$ score.
We use RaAug as the default data augmentation strategy
for our approach when compared with other methods for in-dataset evaluation.

\noindent\textbf{CAM visualizations.}
We visualize the Class Activation Mapping (CAM)
of our approach in Fig.~\ref{fig:vis-cam}. 
The CAM highlights an area inside the forgery region.

\subsection{Cross-dataset Ablation Study}

In this section, we conduct the same experiments across datasets.
We train our models on FF++ (HQ) dataset and evaluate on three datasets: DFD, DFDC-P and Celeb-DF. 
We use DFDC\_selim data augmentation and balance weight $\alpha=1$ as default. 

\begin{table}[t]
    \centering
    \setlength{\tabcolsep}{6pt}
        \small
            \renewcommand{\arraystretch}{1.3}
    \caption{Cross-dataset comparison between consistent representation learning
    to Xception, and Xception+ (Xception trained with
    DFDC\_selim data augmentation).
    Avg refers to average of AUCs.
    Our proposed method outperforms than Xception+ under Avg.
    $^\dagger$ Our reproduced result.}
    \begin{tabular}{l|ccc|c}
        \shline
        Method 
        & DFD & DFDC-P & Celeb-DF & Avg \\
        \shline
        Xception~\cite{chollet2017xception} & ${87.860}$ & $64.724^\dagger$ & $73.040$ & $75.208$\\ 
        Xception+ & $\textbf{95.128}$ & $70.725$ & $70.010$ & $78.621$\\ 
        \hline
        Ours & $94.090$ & $\textbf{72.410}$ & $\textbf{75.718}$ & $\textbf{80.739}$ \\ 
        \shline
    \end{tabular}
    \label{tab:cross-results-base}
\end{table}

\begin{figure*}[t]
  \centering
  \begin{subfigure}{0.49\linewidth}
    \includegraphics[width=\textwidth]{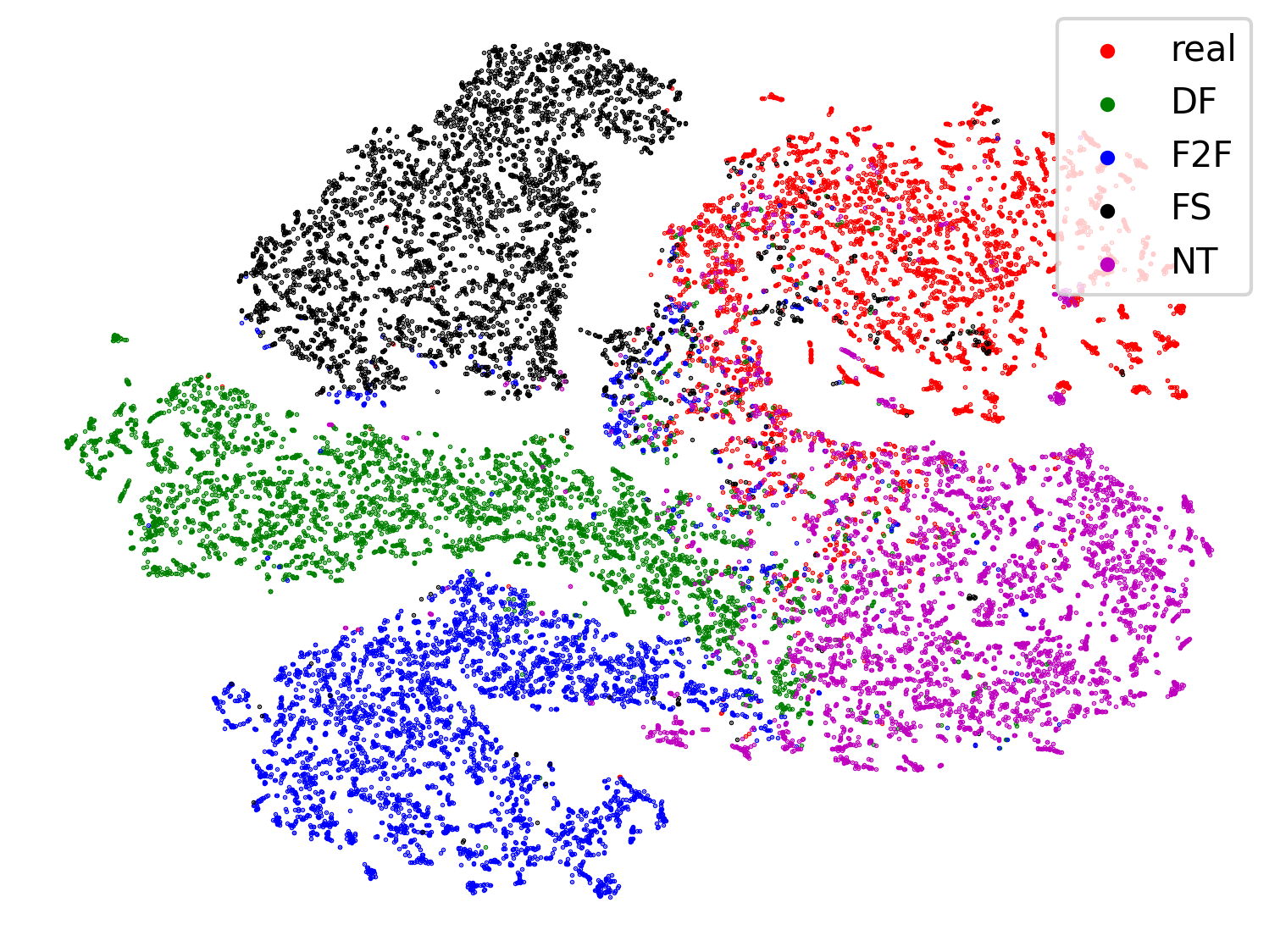}
    \caption{Baseline on FF++(Test) \\ AUC = $98.00$.}
    \label{fig:vis-tsne-a}
  \end{subfigure}
  \hfill
  \begin{subfigure}{0.49\linewidth}
    \includegraphics[width=\textwidth]{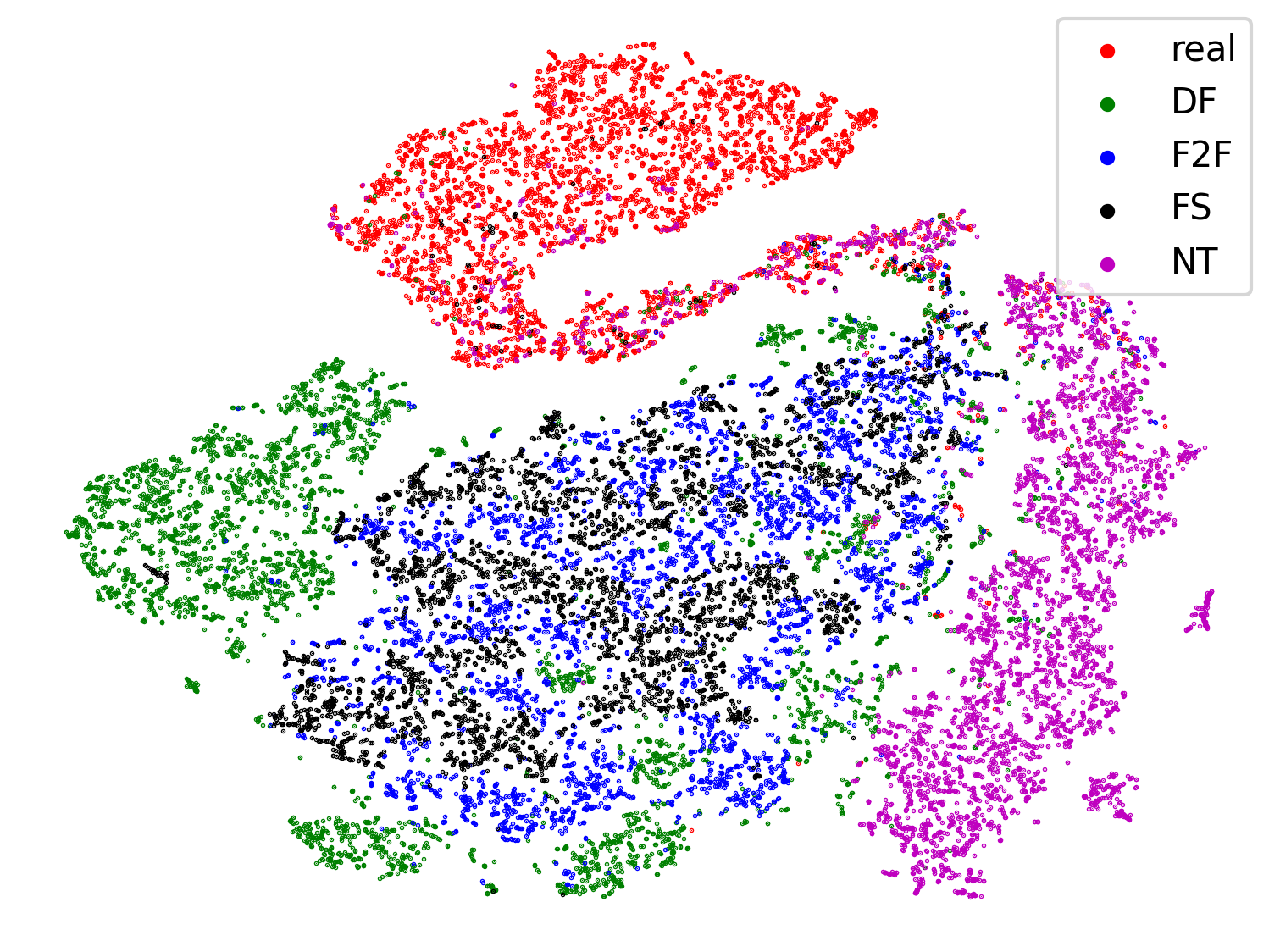}
    \caption{CORE on FF++(Test) \\ AUC = $99.04$.}
    \label{fig:vis-tsne-b}
  \end{subfigure}
  \caption{An illustration of the feature distribution of baseline model (trained with DFDC\_selim data augmentation) and CORE on the intra-domain dataset (FF++) via t-SNE. }
  \label{fig:vis-tsne}
\end{figure*}

\begin{table}[t]
    \centering
    \setlength{\tabcolsep}{8pt}
        \small
            \renewcommand{\arraystretch}{1.3}
    \caption{Cross-dataset ablation study on different penalties
    for consistency loss. 
    Cosine penalty performs the best under the average AUC.}
    \begin{tabular}{l|ccc|c}
        \shline
        Penalty 
        & DFD & DFDC-P & Celeb-DF & Avg\\
        \shline
        $-$  & $95.128$ & $70.725$  & $70.010$ & $78.621$\\ 
        L$1$ & $\textbf{95.621}$ & $67.762$ & $69.853$ & $77.745$  \\ 
        L$2$ & $95.131$ & $67.334$ &  $75.374$ & $76.976$\\ 
        Cos. & $94.090$ & $\textbf{72.410}$ & $\textbf{75.718}$ & $\textbf{80.739}$\\ 
        \shline 
    \end{tabular}
    
    \label{tab:cross-ablation-loss}
\end{table}

\begin{table}[t]
    \centering
    \setlength{\tabcolsep}{10pt}
        \small
            \renewcommand{\arraystretch}{1.3}
    \caption{Cross-dataset ablation study on balance weight $\alpha$. 
    Model with $\alpha=100$ performs the best under the average AUC.}
     \begin{tabular}{l|ccc|c}
        \shline
        $\alpha$ 
        & DFD & DFDC-P & Celeb-DF & Avg \\
        \shline
        $0$  & $95.128$ & $70.725$  & $70.010$ & $78.621$ \\ 
        $1$ & $94.090$ & $72.410$ &  $75.718$ & $80.739$ \\
        $2$ & $\textbf{96.137}$ & $74.834$ &  ${77.267}$ & ${82.746}$ \\
        $5$ & $95.981$ & $72.158$ & $77.091$ & $81.743$ \\
        $10$ & $93.489$ & $\textbf{76.264}$ &  $75.737$ & $81.830$\\
        $100$ & $93.736$ & $75.741$ & $\textbf{79.448}$ & $\textbf{82.975}$\\
        \shline
    \end{tabular}
    \label{tab:cross-ablation-weight}
\end{table}

\begin{table}[t]
    \centering
    \setlength{\tabcolsep}{5pt}
        \small
            \renewcommand{\arraystretch}{1.3}
    \caption{Cross-dataset ablation study on data augmentation. 
    Our approach performs consistently better
    than baseline on all augmentations.
    Ours with DFDC\_selim performs the best. 
    $^\dagger$ Our reproduced result.}
    \begin{tabular}{l|ccc|c}
        \shline
        Data Aug.  &  DFD & DFDC-P & Celeb-DF & Avg \\
        \shline
        None & ${87.860}$ & $64.724^\dagger$ & $73.040$ & $75.208$\\ 
        \multirow{1}{*}{RaAug} 
         & $\textbf{96.188}$ & $67.186$ & $66.983$ & $76.786$\\ 
        \multirow{1}{*}{DFDC\_selim} 
         & $93.736$ & $\textbf{75.741}$  & $\textbf{79.448}$ & $\textbf{82.975}$\\ 
        \shline
    \end{tabular}
    \label{tab:cross-results-dataaug}
\end{table}

\noindent\textbf{Effectiveness of consistent representation learning.} 
In this section, Xception+ refers to Xception trained with DFDC\_selim data augmentation.
We select different data augmentation from the in-dataset experiments because different data augmentations have different effects on generalization performance.
Ours adopts the same data augmentation as Xception+.
As shown in Tab.~\ref{tab:cross-results-base}, 
Xception+ outperforms than Xception.
We used AUC as the default metric in cross-dataset ablation study experiments.
Our proposed method outperforms Xception+ under the average AUC by $2.118\%$.
In particular, our proposed method achieves a $5.708$ AUC gain under Celeb-DF dataset.

\noindent\textbf{Ablation study on consistency loss.} 
As shown in Tab.~\ref{tab:cross-ablation-loss}, different penalties show a similar effect as the in-dataset experiments.
Cosine penalty performs better than L1 and L2 penalties under the average AUC.
This proves that consistency representation learning also works across datasets.

\begin{table*}[t]
    \centering
    \setlength{\tabcolsep}{12pt}
        \small
            \renewcommand{\arraystretch}{1.3}
    \caption{In-dataset SOTA comparisons on FF++ dataset. Our proposed approach performs the best under RAW and HQ quality settings. $^\ddagger$ we report the results
    provided in FF++ paper~\cite{rossler2019faceforensics++}.
    }
    \begin{tabular}{l|l|cc|cc|cc}
        \shline
        \multirow{2}{*}{Method} & \multirow{2}{*}{Reference} & \multicolumn{2}{c|}{RAW}
         & \multicolumn{2}{c|}{HQ} & \multicolumn{2}{c}{LQ} \\
         \cline{3-8}
         & & Acc. & AUC & Acc. & AUC & Acc. & AUC \\
        \shline
        Steg. Features~\cite{fridrich2012rich}$^\ddagger$
        & IEEE TIFS 2012 & $97.63$ & $-$ & $70.97$ & $-$ & $55.98$ & $-$ \\ 
        C-Conv~\cite{bayar2016deep}$^\ddagger$ &
        IH\&MMSec 2016  & $98.74$ & $-$ & $82.97$ & $-$ & $66.84$ & $-$ \\ 
        LD-CNN~\cite{cozzolino2017recasting}$^\ddagger$ &
        IH\&MMSec 2017 & $98.57$ & $-$ & $78.45$ & $-$ & $58.69$ & $-$ \\ 
        CP-CNN~\cite{rahmouni2017distinguishing}$^\ddagger$ &
        WIFS 2017 & $97.03$ & $-$ & $79.08$ & $-$ & $61.18$ & $-$ \\ 
        Xception~\cite{chollet2017xception}$^\ddagger$ &
        CVPR 2017 & $99.26$ & $-$ & $95.73$ & $-$ & $86.86$ & $-$ \\
        MesoNet~\cite{afchar2018mesonet}$^\ddagger$ &
        WIFS 2018 & $95.23$ & $-$ & $83.10$ & $-$ & $70.47$ & $-$ \\ 
        Two-branch RN~\cite{masi2020two} & ECCV 2020 & $-$ & $-$ & $96.43$ & $88.70$ & $86.34$ & $86.59$ \\
        F3-Net~\cite{qian2020thinking} & 
        ECCV 2020 & $99.95$ & $99.80$ & $97.52$ & $98.10$ & $90.43$ & $93.30$ \\
        DeepfakeUCL~\cite{fung2021deepfakeucl} & IJCNN 2021 & $-$ & $-$ & $-$ & $93.00$ & $-$ & $-$ \\
        RFAM~\cite{chen2021local} & AAAI 2021 & $99.87$ & $99.92$ & $97.59$ & $99.46$ & $\mathbf{91.47}$ & $\mathbf{95.21}$ \\
        SPSL~\cite{liu2021spatial} & CVPR 2021 & $-$ & $-$ & $91.50$ & $95.32$ & $81.57$ & $82.82$ \\
        Multi-attention~\cite{zhao2021multi} & CVPR 2021 & $-$ & $-$ & $97.60$ & $99.29$ & $88.69$ & $90.40$ \\
        \rowcolor{Gray} CORE & & $\mathbf{99.97}$ & $\mathbf{100.00}$ & $\mathbf{97.61}$ & $\mathbf{99.66}$ & 87.99 & 90.61 \\
         \shline
    \end{tabular}
    \label{tab:results-ffpp}
\end{table*}

\begin{table}[t]
    \centering
    \setlength{\tabcolsep}{2.7pt}
        \small
            \renewcommand{\arraystretch}{1.3}
    \caption{In-dataset SOTA comparisons on CELEB-DF dataset. Our proposed approach performs the best. $^\dagger$ we report the results provided in ~\cite{wang2021representative}.
    }
    \begin{tabular}{l|cccc}
        \shline
        Method & Acc. & AUC & TDR@$0.1\%$ & TDR@$0.01\%$ \\
        \shline
        Hu et al.~\cite{hu2021detecting} & $80.74$ & $87.00$ & $-$ & $-$ \\
        FakeCatcher~\cite{ciftci2020fakecatcher} & $91.50$ & $-$ & $-$ & $-$ \\
        XcepTemporal~\cite{chintha2020recurrent} & $97.83$ & $-$ & $-$ & $-$ \\
        Xception~\cite{chollet2017xception}$^\dagger$ & $-$ & $99.85$ & $89.11$ & $84.22$ \\
        RE~\cite{zhong2020random}$^\dagger$ & $-$ & $99.84$ & $84.05$ & $76.63$ \\
        AE~\cite{wei2017object}$^\dagger$ & $-$ & $99.89$ & $88.11$ & $85.20$ \\
        Patch~\cite{chai2020makes}$^\dagger$  & $-$ & $99.96$ & $91.83$ & $86.16$  \\
        RFM-X~\cite{wang2021representative}$^\dagger$ & $-$ & $99.94$ & $93.88$ & $87.08$  \\
        RFM-Patch~\cite{wang2021representative}$^\dagger$ & $-$ & $\mathbf{99.97}$ & $93.44$ & $89.58$  \\
        \rowcolor{Gray} CORE & $\mathbf{99.17}$ & $\mathbf{99.97}$ & $\mathbf{95.58}$ & $\mathbf{93.98}$ \\
         \shline
    \end{tabular}
    \label{tab:results-celebdf}
\end{table}

\begin{table}[t]
    \centering
    \setlength{\tabcolsep}{8.0pt}
        \small
            \renewcommand{\arraystretch}{1.3}
    \caption{In-dataset SOTA comparisons on DFFD dataset. Our proposed method performs the best.
    $^\dagger$ we report the results provided in ~\cite{dang2020detection}.
    }
    \begin{tabular}{l|ccc}
        \shline
        Method & AUC & TDR$_{0.1\%}$ & TDR$_{0.01\%}$ \\
        \shline
        Xception~\cite{chollet2017xception}$^\dagger$ & $99.61$ & $85.26$ & $77.42$ \\
        Reg-Xception~\cite{dang2020detection}$^\dagger$ & $99.64$ & $90.78$ & $83.83$ \\
        RFM-Xception~\cite{wang2021representative} & $99.97$ & $98.35$ & $95.50$ \\
        \rowcolor{Gray} CORE & $\mathbf{99.99}$ & $\mathbf{99.23}$ & $\mathbf{98.15}$  \\
         \shline
    \end{tabular}
    \label{tab:results-dffd}
\end{table}

\begin{table}[t]
    \centering
    \setlength{\tabcolsep}{9pt}
        \small
            \renewcommand{\arraystretch}{1.3}
    \caption{In-dataset SOTA comparisons on DFDC-P dataset. Our proposed approach achieve similar performance to the SOTA method. 
    }
    \begin{tabular}{l|cc}
        \shline
        Method & Acc. & AUC  \\
        \shline
        Tolosanaet al.~\cite{tolosana2021deepfakes} & $-$ & $91.10$ \\
        S-MIL-T~\cite{li2020sharp} & $-$ & $85.11$ \\
        LSC~\cite{zhao2021learning} & $-$ & $\textbf{94.38}$  \\
        \rowcolor{Gray} CORE & $84.38$ & $92.31$  \\
         \shline
    \end{tabular}
    \label{tab:results-dfdc-p}
\end{table}

\noindent\textbf{Ablation study on balance weight $\alpha$.}
As in in-dataset experiments, we explore some different balance weight $\alpha$.
As illustrated in Tab.~\ref{tab:cross-ablation-weight}, $\alpha=100$ performs better under the average AUC and most datasets.
The best balance weight differs between the in-dataset and cross-dataset experiments.
This shows that stronger consistency constraints work better in cross-domain experiments.

\noindent\textbf{Ablation study on data augmentation.} 
We also conduct experiments on different data
augmentation strategies:
proposed RaAug and DFDC\_selim (see Sec.~\ref{sec:implementation}
for details) under $\alpha=100$.
As shown in Tab.~\ref{tab:cross-results-dataaug},
data augmentation strategy plays an important role in CORE framework, which
improves model generalization by a big gap.
DFDC\_selim achieves big gain in all three datasets.
This shows that more complex data augmentation performs better in cross-domain experiments.

\begin{table*}[htbp]
    \centering
    \setlength{\tabcolsep}{9pt}
        \small
            \renewcommand{\arraystretch}{1.3}
    \caption{Cross-dataset evaluation results on DFD, DFDC-P and Celeb-DF in terms of AUC. 
    Our proposed method achieves highest AUC on DFD dataset
    and second highest AUC on DFDC-P and Celeb-DF dataset.
    }
    \begin{tabular}{l|l|ll|ccc}
        \shline 
        Method & Reference & Backbone & Train Set &  DFD & DFDC-P & Celeb-DF \\
        \shline 
        Xception~\cite{chollet2017xception} & CVPR 2017 & Xception & FF++
        & $87.86$ & $-$ &  $73.04$ \\ 
        DSP-FWA~\cite{li2018exposing} & CVPRW 2019 & ResNet-50 & FF++
        & $-$ & $-$ & $69.30$ \\
        EfficientNet~\cite{tan2019efficientnet} & ICML 2019 & EfficientNet-B4
        & FF++ & $-$ & $-$ & $64.29$ \\
        Face X-ray~\cite{li2020face} & CVPR 2020 & HRNet & BI (private dataset)
        & \underline{$93.47$} & $71.15$  & $74.76$ \\
        Two-branch RN~\cite{masi2020two} & ECCV 2020 & LSTM & FF++
        & $-$ & $-$  & $73.41$ \\
        F3-Net~\cite{qian2020thinking} & ECCV 2020 & Xception & FF++
        & $-$ & $-$  & $76.88$ \\
        DeepfakeUCL~\cite{fung2021deepfakeucl} & IJCNN 2021 & Xception & FF++ &  $-$ & $-$ & $56.80$ \\
        Local-relation~\cite{chen2021local} & AAAI 2021 & Xception & FF++
        & $89.24$ & $\textbf{76.53}$  & $78.26$ \\
        HFF~\cite{luo2021generalizing} & CVPR 2021 & Xception (modified)
        & FF++ & {$91.90$} & $-$  & {$79.4$} \\
        Multi-attention~\cite{zhao2021multi} & CVPR 2021
        & EfficientNet-B4 & FF++ & $-$  & $-$  & $67.44$ \\
        SPSL~\cite{liu2021spatial} & CVPR 2021 & Xception & FF++ & $-$ & $-$ & $76.88$ \\
        LSC~\cite{zhao2021learning} & ICCV 2021 & ResNet-$34$ & FF++ (real data)
        & $-$ & $74.37$  & $\mathbf{81.80}$ \\
        \rowcolor{Gray} CORE & & Xception & FF++
        & $\mathbf{93.74}$  & \underline{$75.74$}  & \underline{$79.45$} \\
        \shline
    \end{tabular}
    \label{tab:results-cross}
\end{table*}

\noindent\textbf{t-SNE visualizations.}
We visualize the feature distribution of the test part of FF++ dataset via t-SNE~\cite{van2008visualizing}.
FTCN~\cite{zheng2021exploring} observes that the CNN model can easily extract the unique artifacts of different manipulated methods, even if training with all manipulated data as one class.
As illustrated in Fig.~\ref{fig:vis-tsne-a} and Fig.~\ref{fig:vis-tsne-b}, the baseline model can separate four different forgery methods with a gap than CORE. CORE doesn't distinguish the type of forgery algorithm to some degree.
This indicates that CORE extracts the essential features for forgery detection instead of manipulated artifacts.
Tab.~\ref{tab:cross-results-dataaug} shows that CORE achieve $6.408$ AUC gains under Celeb-DF and $11.017$ AUC gains under DFDC-P dataset.

\subsection{In-Dataset Comparison to Other Methods}

In this section, we compare our method with previous forgery
detection methods under four datasets: FF++, Celeb-DF, DFFD and DFDC-P.

\noindent\textbf{Evaluation on FaceForensics++.}
The comparisons are shown in Tab.~\ref{tab:results-ffpp}. 
Our approach outperforms all the other methods under RAW and
HQ quality settings.
Compared with Xception~\cite{rossler2019faceforensics++},
which uses the same backbone network, our method performs better
under all quality settings.
Our method performs better than DeepfakeUCL~\cite{fung2021deepfakeucl},
which adopts the contrastive learning method.
For LQ setting, our method performs inferior to RFAM
and F3-Net, similar to Multi-attention.
RFAM and F3-Net encode features in the frequency domain,
which has proven to be effective for LQ images.
We do not adopt frequency-domain processing modules
and this might lead to the superior performance
of their approaches to ours.

\noindent\textbf{Evaluation on Celeb-DF.}
As shown in Tab.~\ref{tab:results-celebdf}, our method
performs the best, 
achieving $91.17$ in ACC, $99.97$ in AUC, 
$95.58$ in TDR$_{0.1\%}$,
and $93.98$ in TDR$_{0.01\%}$. 
Compared with the previous state-of-the-art method Patch,
our approach improves TDR$_{0.1\%}$ and TDR$_{0.01\%}$
with a large margin.

\noindent\textbf{Evaluation on DFFD.} 
Tab.~\ref{tab:results-dffd} shows that our proposed method achieves the best performance. Compared to the previous state-of-the-art method RFM~\cite{wang2021representative}, our method yields improvements of $0.02$ AUC, $0.88$ TDR$_{0.1\%}$,
and $2.65$ TDR$_{0.01\%}$ under the same training and evaluation data.

\noindent\textbf{Evaluation on DFDC-P.} 
Tab.~\ref{tab:results-dfdc-p} show that our proposed approach achieve similar performance to the state-of-the-art method.
DFDC-P dataset contains considerable low-quality videos. 
Models can't performs well without frequency information.
That a very simple method get the second place would also proves the effectiveness of consistent representation learning.

\subsection{Cross-Dataset Comparison to Other Methods}

We conduct cross-dataset evaluation to demonstrate the generalization of our proposed method CORE.
We trained our model using FF++ (HQ) dataset and evaluate on DFD, DFDC-P, and Celeb-DF.
The results in Tab.~\ref{tab:results-cross} show that our proposed method can obtain state-of-the-art performance under a sample framework with a vanilla backbone.
Our proposed achieve a big gain than Xception~\cite{chollet2017xception} in all three cross datasets.
As for DFDC-P and Celeb-DF datasets, our proposed method obtains second performance and is similar to the state-of-the-art performance.

\section{Conclusion}

In this paper, we introduce a simple yet effective framework,
Consistent Representation Learning (CORE).
CORE employs different augmentations for the input 
and explicitly constrains the consistency of different representations.
The proposed framework is flexibly integrated with almost any other method.
Extensive quantitative and qualitative comparisons (in-dataset and cross-dataset) show that CORE performs
favorably against recent state-of-the-art
face forgery detection approaches.


{\small
\bibliographystyle{ieee_fullname}
\bibliography{main}
}

\end{document}